# IMPROVING THE CAPABILITIES OF LARGE LANGUAGE MODEL BASED MARKETING ANALYTICS COPILOTS WITH SEMANTIC SEARCH AND FINE-TUNING

Yilin Gao[1], Sai Kumar Arava[2], Yancheng Li[2], and James W Snyder Jr[2]

[1]Department of Quantitative and Computational Biology, University of Southern California, Los Angeles, California, USA
[2]Adobe Inc., San Jose, California, USA
snyderjr@adobe.com

## ABSTRACT

*Artificial intelligence (AI) is widely deployed to solve problems related to marketing attribution and budget optimization. However, AI models can be quite complex, and it can be difficult to understand model workings and insights without extensive implementation teams. In principle, recently developed large language models (LLMs), like GPT-4, can be deployed to provide marketing insights, reducing the time and effort required to make critical decisions. In practice, there are substantial challenges that need to be overcome to reliably use such models. We focus on domain-specific question-answering, SQL generation needed for data retrieval, and tabular analysis and show how a combination of semantic search, prompt engineering, and fine-tuning can be applied to dramatically improve the ability of LLMs to execute these tasks accurately. We compare both proprietary models, like GPT-4, and open-source models, like Llama-2-70b, as well as various embedding methods. These models are tested on sample use cases specific to marketing mix modeling and attribution.*

## 1. INTRODUCTION

Marketing is an important function of many businesses. Historically, marketing was done primarily through a limited number of channels, such as print, radio, and linear TV advertising, but as the internet became more widely used, digital marketing increased dramatically. One of the benefits of digital marketing was the trove of additional data that companies could acquire related to the efficacy of their marketing efforts. However, drawing insights from this data requires significant effort from analysts, often utilizing a variety of different marketing software solutions. Machine learning based marketing software further complicates this effort because, while providing more sophisticated and accurate insights, it can be difficult to understand how models work. To address these difficulties, enterprise software companies often have extensive implementation teams that configure applications and explain insights in detail. This situation is not optimal because it makes deployments more expensive, and it can take a nontrivial amount of time to get questions answered as they arise. As such, reducing the need for human involvement has significant value.

Large language models (LLMs) are essentially large transformer-based machine learning models that mine statistical relationships in text data to generate additional text. Although LLMs had been the topic of research for many years [1-13], the release of ChatGPT in early 2023 significantly increased interest in chatbot assistants as a useful supplement to more traditional forms of information lookup. Interest in LLMs extended beyond individuals, as many companies released LLM based personal assistants [14-15]. LLMs would also be useful in marketing analytics software. However, LLMs are not equally effective at every task. LLMs struggle with SQL generation and tabular analysis, particularly involving numbers, because they were not trained specifically for these tasks. Even for use cases that focus on natural language, LLMs often suffer from the well documented hallucination problem [16-18], especially in cases where lengthy or domain-specific context is important. Since reliably answering domain-specific questions, writing SQL code to retrieve data, and analyzing tables is critical in a marketing analytics tool, these are important limitations. Unlike use cases involving image generation or product suggestions, returning incorrect information in an analytics tool can be very detrimental to a business, and LLMs cannot be deployed accurately without improvements and guardrails.

Fortunately, there are various ways to mitigate these deficiencies, which are collectively called refinement methods. Strategies, such as semantic search, have been used to embed documents as a numerical vector and match a query to relevant information, which can then be used as context within an LLM. Detailed prompt engineering is even more commonly used to provide LLMs with context to accurately answer a variety of questions, but it is only useful in situations where more limited context is sufficient. These methods are especially effective for tasks that LLMs usually handle well, but they are generally insufficient for use cases involving numerical and tabular analysis. Fine-tuning, which involves updating the weights of an LLM based on further training on a limited amount of additional data, is another approach used to improve the functionality of LLMs on novel tasks. It is particularly effective at helping LLMs generalize to entirely new tasks, such as numerical and tabular analysis.

In this paper, we demonstrate how these different refinement methods can be used to do domain-specific question-answering, SQL code generation, and tabular analysis, which are necessary to create an effective marketing analytics assistant. Specifically, we will focus on marketing mix modeling and attribution use cases, which are used to help companies understand how to efficiently allocate marketing budget across various media channels and strategies. For context, mix modeling allows business to understand the contribution of marketing and various internal and external factors on their sales (or other business objective) at an aggregate level, typically a daily or weekly time series [19-21]. Attribution, in contrast, helps marketers understand the detailed interaction between digital media marketing and some conversion, such as sales [22-30]. While attribution is much more granular and precise, it requires data to be stitched to individual users, which is only possible for some digital media marketing and is becoming increasingly difficult with the removal of support for third party cookies. In contrast, mix modeling lacks these constraints, but it has more limited resolution. As such, the two methods complement each other and can be combined for a more holistic marketing analytics tool. We focus on how various LLMs perform, with and without refinement, on a variety of tasks that are important to deploying a marketing analytics copilot.

This work contributes to the literature in two respects. First, it demonstrates how LLMs can be used in analytics tools, with a specific focus on marketing analytics tools. Second, it shows the relative performance of different widely available LLMs, with and without refinement, which is useful information for anyone working on LLM based applications. As such, this work is broadly useful to individuals that both work in marketing software as well as in applied machine learning.

## 2. RELATED WORK

### 2.1. Semantic Search

Semantic search is an advanced information retrieval method that understands the intent and contextual meaning of a query, rather than just the key words. Modern search engines now employ these principles, integrating artificial intelligence to enhance user experience by delivering contextually relevant results. Semantic search encompasses three primary steps:

1. Creation of a custom knowledge base, wherein texts are converted into numerical vectors using an embedding model
2. Retrieval of pertinent information from this knowledge base by juxtaposing it with the user's query
3. Deployment of an LLM's capabilities to craft a response based on the amalgamated information from the knowledge base and the user's query

Semantic search has been a topic of interest for many years, aiming to improve the precision and relevance of search results by understanding the intent and contextual meaning of terms within the search query.

Traditional approaches to semantic search often relied on structured knowledge graphs, ontologies, and semantic web technologies. Prior to the development of LLMs, semantic search had been leveraged for various business applications. For example, the semantic search engine was developed as a web service that stores semantic information about web resources and can solve complicated queries, considering the context and target of the web resource [31]. This approach can be particularly beneficial for discovering information about commercial products and services. Further research demonstrated how semantic web technology could be applied to other fields and use cases relevant to enterprises [32]. One issue with early semantic web technology was the manual effort required to build up knowledge graphs and ontologies, but frameworks were eventually developed to partially automate the construction and maintenance of ontologies [33]. Semantic frameworks were developed for ecommerce search engine optimization, emphasizing the importance of structured data and semantic annotations [34].

Even with the previously mentioned automation efforts, building knowledge graphs and ontologies for semantic search technology was still effort intensive. This changed with the advent of deep learning. There has been a shift towards using neural network-based models for semantic search. Word embeddings, such as Word2Vec and GloVe, were among the first to represent words in continuous vector spaces, capturing semantic relationships between terms. These embeddings later enhanced search algorithms by considering the semantic similarity between query and document terms. Furthermore, the emergence of LLMs, like GPT-4, has significantly advanced the field of natural language processing. For instance, Chameleon, an AI system that augments LLMs with ready-to-use compositional reasoning modules, showcased the effectiveness of LLMs in complex reasoning tasks [35]. ChatGPT was shown to be useful in healthcare education, research, and practice, specifically in scientific writing, efficient analysis of massive datasets, and improved health literacy [36]. Finally, LLMs demonstrated potential in more realistic semantic parsing tasks, even when utilizing larger vocabularies [37].

In summary, while traditional methods of semantic search relied heavily on structured data and ontologies, the rise of LLMs and advancements in semantic technologies have introduced new possibilities for more flexible and context-aware search mechanisms, especially in business applications. Our work aims to build upon these foundational works and explore novel applications of LLMs in the realm of semantic search and marketing analytics.

## 2.2. Prompt Engineering

Prompt engineering involves designing specific inputs or "prompts" to steer language models. Since LLMs are input-sensitive, well-crafted prompts can elicit desired outputs without changing the model's internal structure. This technique is especially useful with models, like GPT-4, enabling users to influence model responses without retraining [38]. While we do employ prompt engineering in this paper to provide further context to various pretrained and fine-tuned LLMs, it is not our primary focus, and its discussion will be more limited. The key point to make about prompt engineering is that it is a useful way to improve performance of LLMs on specific tasks, with or without other refinement methods, and it is often the easiest refinement method to try.

## 2.3. Fine-tuning

Fine-tuning, often used interchangeably with transfer learning, is the practice of further training a pretrained model on a specific dataset to tailor it for a particular task. It capitalizes on the model's pre-existing knowledge, enhancing its performance for specialized tasks and often resulting in better accuracy, even with limited data. This approach is prevalent in deep learning, notably in image and natural language processing (NLP) [39]. LLMs are primarily designed to generate new natural language text, based on prior text. While this is adequate for answering questions based on text documents, this does not necessarily work for generating SQL code or summarizing tables, despite this data's potential inclusion in the LLMs training data. This is a serious shortcoming because marketing analytics tools need to answer questions based on tabular data, usually stored in a database.

Although coding in SQL may seem similar to language, there are some key differences that make generating SQL code challenging. LLMs struggle with queries requiring multiple steps or heavy contextual reliance, since this demands a deep semantic grasp of the query's intent as opposed to mere pattern recognition [40]. Related to this, the inherent ambiguity in natural language can lead to SQL queries that might be syntactically correct but semantically off-target [41]. Moreover, generating intricate SQL queries, especially those with multiple joins or conditions, is challenging, with accuracy decreasing as query complexity increases [42]. Lastly, LLMs cannot directly interface with specific database schemas, making it difficult to produce SQL commands tailored to a database's unique structure [43]. Tabular data analysis is even more problematic. LLMs, designed for linear text, struggle with tables' spatial information encoding. Translating interconnected rows and columns into coherent language is complex. In addition, LLMs are built on text relationships and do not understand numerical relationships. This means that they are unreliable in terms of handling numerical data, especially more complex numbers like floats or doubles.

Despite these challenges, prior research has made significant improvements in these areas. Specialized LLMs have shown proficiency in code generation, especially in SQL for databases. SQL-PaLM, an LLM-based Text-to-SQL model, set new benchmarks in both in-context learning and fine-tuning contexts [44]. DBVinci, based on OpenAI's GPT-3.5 Text-davinci-003 engine, can execute a variety of tasks, including text-to-SQL and inserting code completions. The model breaks complicated SQL queries into simpler steps, which are described with natural language, and achieves competitive results without the need for fine-tuning on large-scale annotated datasets [45]. ZeroNL2SQL uses pretrained language models for database schema alignment and LLMs for complex language reasoning [46]. Finally, TAPEX, a pioneering approach that emphasizes table pretraining, involves training a neural network on a query and table to produce tabular output. This training data is uniquely produced by automatically generating SQL queries and their corresponding outputs, as determined after executing the query. Notably, TAPEX has established new benchmarks in several related tasks [47], particularly its top-ranking performance on the WikiSQL dataset [43], underscoring the potential of the method they proposed.

Ultimately, automatically generating SQL code is not very useful without the ability to analyze the corresponding results, so additional work has been done to develop models that can interact with tables using natural language. TableGPT is a software package, based on fine-tuned LLMs, that enables users to query and manipulate tables using natural language and various functional commands [48]. Chameleon also implements modules, with the use of LLMs, on table verbalization and knowledge retrieval [35].

While recent research has made many breakthroughs in the field of SQL code generation and tabular data analysis, these tasks remain challenging due to their complexity and ambiguity in syntax and structures. To date, these methods are not reliable enough to be used in a production application without a human in the loop. In this paper, we show how fine-tuning can be used to improve SQL query generation accuracy and tabular data analysis.

## 3. METHOD

### 3.1. Domain-specific Question-answering

We apply semantic search techniques to answer domain-specific questions based on a wide variety of documents, aiming to create a question-answering system that can be effectively used in marketing analytics tools. This semantic search process is illustrated in Figure 1, with details provided below.

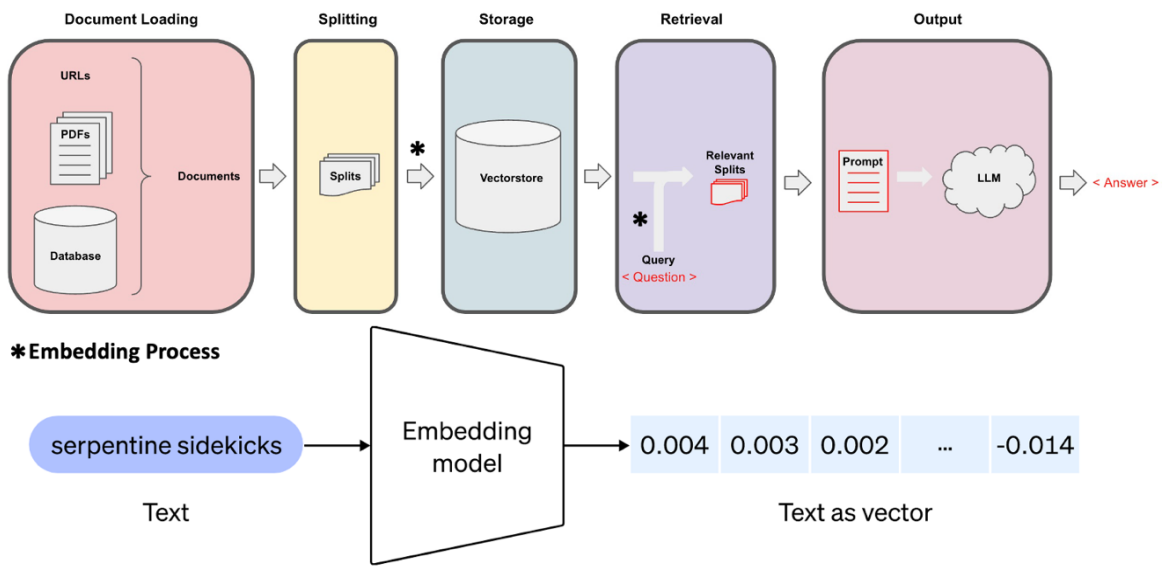

Figure 1. An illustration of the semantic search procedure.

**Knowledge base construction:** We developed a marketing analytics knowledge base by curating a dataset from 124 unique URLs, mainly from a large company's public domains and some from internal wikipages that contain publicly shared information. Using web crawling, we extracted textual data, tokenized it, and segmented it into chunks of up to 500 tokens. This resulted in a total of 1,264 document chunks. These segments were then converted into embedding vectors using advanced algorithms. The final step involved integrating these vectors into a database, creating our knowledge base. We used OpenAI's FAISS (Facebook AI Similarity Search) for storing and retrieving these embedding vectors.

**Text embedding models:** To evaluate the efficacy of various text embedding models and assess their influence on semantic search tasks, we employed OpenAI's text-embedding-ada-002 alongside open-sourced INSTRUCTOR-LARGE and INSTRUCTOR-XL from the INSTRUCTOR model family [49]. At the time, these were the best performing text embedding models on the Huggingface leaderboard. To compare the capability of different text embedding models, we used OpenAI's GPT-4 as the LLM for the process of semantic search.

**LLMs used to construct responses:** To enhance our semantic search's question-answering capability, we employed several top-tier language models, including GPT-4. We also utilized open-sourced models, Falcon-40B [50] and Llama-2-70b [51] which are amongst the most powerful open-sourced models. These specific models were chosen because they are the largest LLMs with some of the best published performance characteristics. We felt they were the only ones likely to have comparable performance to GPT-4, which is widely considered the gold standard for LLMs. In general, we expect models to perform better as the number of parameters increases, but the specifics of the training data may have a nontrivial impact on model performance as well. For example, some LLMs may have been trained on more coding text, allowing them to perform better in these cases. It is difficult to know how the training data will impact performance in advance, unless performance metrics were previously published against a variety of similarly sized LLMs. For comparing these LLMs, we employed OpenAI's text-embedding-ada-002 as our semantic search's text embedding model.

**Evaluations leveraging GPT-4:** In our endeavor to find the best LLM and text embedding model for semantic search in marketing analytics documents, we adopted an approach inspired by G-Eval [52]. Our analysis utilized GPT-4 to evaluate LLM-generated responses to a set of user queries. Our evaluation considered accuracy, relevance, thoroughness, clarity, and conciseness, scoring responses on a 1-5 scale. This allowed us to compare the performance of different LLMs and text embedding models.

We designed six queries, each related to unique marketing analytics scenarios, and generated multiple responses from various LLM and text embedding combinations. Using GPT-4, we compared these responses to human-crafted answers. Ideally, multiple unbiased experts would evaluate these, but due to costs, we used AI. Other quantitative methods, like BLEU scores, are not as effective for use cases where there are many different reasonable answers. By contrasting AI-generated responses with human benchmarks, we gauged their accuracy and relevance, aiming to identify the top-performing models.

## 3.2. SQL Query Generation

Fine-tuning is employed to teach LLMs to generate SQL code, which many LLMs can do to some extent but with substantial error. Prompt engineering is also used to provide additional context to all models and is the only refinement method applied to GPT-4, since it is proprietary.

**Data:** We used the b-mc2/sql-create-context dataset from Huggingface, containing 78,577 examples. Each example pairs a natural language question with an SQL context describing the data schema and an SQL answer. An example is provided in Table 1. We used this dataset because it is large, comprehensive, and publicly available. It is roughly the same size as the commonly used WikiSQL dataset, which is one of the largest annotated text to SQL datasets. Unlike WikiSQL, it provides the create statements for each table to limit hallucinations, which is a practice that is achievable in a production application. This dataset is not specific to marketing, and the conclusions generated from this study should generalize to any application involving SQL query generation, assuming the tables are somewhat similar to those in the dataset (e.g. they don't have thousands of columns). We reserved 1,000 samples for evaluation and used the remaining 77,577 for model fine-tuning, ensuring a comprehensive training and rigorous performance assessment.

Table 1. Example from b-mc2/sql-create-context dataset.

| | |
|---|---|
| **Question** | How many heads of the departments are older than 56? |
| **Context** | CREATE TABLE head (age INTEGER) |
| **Answer** | SELECT COUNT(*) FROM head WHERE age $>$ 56 |

**LLM approaches:** We compared various methodologies to identify the most effective strategy. In general, we compared the raw pretrained models to fine-tuned models to demonstrate accuracy improvement. However, GPT-4 does not support fine-tuning as its architecture is not public. To demonstrate the top performance possible from a GPT-4 model, we used few-shot learning, which involved providing a few examples in the engineered prompt. We provided GPT-4 with examples from the b-mc2/sql-create-context dataset to guide its SQL query generation. Prompt engineering was also more broadly used to provide each LLM with context for the tables, such as the schema, which is shown in Table 1. Below are the methods we employed:

1. Few-shot learning with GPT-4
2. Pretrained Falcon-40b-4bit
3. Fine-tuned Falcon-40b-4bit
4. Pretrained Llama-2-13b
5. Fine-tuned Llama-2-13b
6. Pretrained Llama-2-70b-4bit

As with the domain-specific question-answering use case, we selected open-sourced LLMs that we thought were likely to perform well against GPT-4. However, we also included Llama-2-13b for this comparison because the Llama-2-70b-4bit model was too large to fine-tune on a single GPU cluster. To still evaluate the improvement possible with fine-tuning on the Llama model family, we used the model with 13 billion parameters in lieu of the 70 billion parameter model. To fine-tune the LLMs to write SQL queries, we trained the models for 5 additional epochs with the training data from the b-mc2/sql-create-context dataset.

### 3.3. Tabular Data Analysis

Analyzing tabular data is not something LLMs were designed to do, and fine-tuning is used to teach the LLMs how to perform a specific tabular analysis task important to a marketing analytics copilot. The specific use case involves explaining why the attribution credit to various touch points is expected to change, based on a variety of contributing and mitigating factors precomputed and contained in a table, after retraining an attribution model on new data. A sample row from the data is shown in Table 2. We set guidelines: values above 5 were contributors, below -5 were mitigators, and others were non-influential. Ideally, the LLM would interpret the table as: lead - display: The absolute change of this channel is -82%, which indicates a decrease in the touch point's credit. The targeting quality is a contributing factor with a score of 63%. The contact frequency is a not a factor with a score of -4%. The ad cannibalization is a mitigating factor with a score of -33%. Prompt engineering was employed to provide each LLM with instructions on how to solve the problem.

Table 2. Example of tabular data understanding task.

| model name | lead |
|---|---|
| channel | display |
| absolute change | -82 |
| targeting quality | 63 |
| contact frequency | -4 |
| ad cannibalization | -33 |

**Data:** We created a large dataset to aid the LLM in understanding and analyzing tabular data. From building attribution models, we had a wide variety of real data in the same format as Table 2, but this data was limited to around 100 examples. We used these examples as a reference within our simulator to generate new data. No specific distribution was enforced because we want the LLM to be able to explain both common and infrequently occurring samples, but we did ensure that the samples were all feasible outputs. We generated approximately 10,000 synthetic samples.

Additionally, we reserved 1,000 samples for evaluation to measure the LLM's ability to derive insights from tables. The size of this dataset was large enough to avoid overfitting without being computationally burdensome. While this task is specific to marketing analytics, it is essentially a specific instantiation of a row-by-row table summarization task, which is common to a variety of other disciplines and should generalize well.

**LLM approaches:** As with SQL query generation, we compared a variety of methods, consisting of various pretrained and fine-tuned models. Since GPT-4 cannot employ fine-tuning, we used GPT-4 with a variety of engineered prompts as well as a third-party pandas agent package. Our prompts explained to GPT-4 how to solve the problem and provided examples in csv, list template, and text template format. The use of csv format means that Table 2 is provided as an example in csv format. This is how it appears in list template format:

**model name**: lead
**channel**: display
**absolute change**: -82
**targeting quality**: 63
**contact frequency**: -4
**ad cannibalization**: -33

In text template format, the example would read "The **model name** is lead, the **channel** is display, the **absolute change** is -82, the **targeting quality** has a value of 63, the **contact frequency** has a value of -4, the **ad cannibalization** has a value of -33." Pandas agent, from Langchain, facilitates interaction with a pandas dataframe by executing Python code generated by the LLM. Here is a list of the methods we employed:

1. CSV with GPT-4
2. List template with GPT-4
3. Text template with GPT-4
4. Langchain pandas agent with GPT-4
5. Pretrained Falcon-40b-4bit
6. Fine-tuned Falcon-40b-4bit
7. Pretrained Llama-2-13b
8. Fine-tuned Llama-2-13b
9. Pretrained Llama-2-70b-4bit

Aside from GPT-4, all other models were provided the table only in csv format. Prompt engineering was used across all models to explain how to solve the problem in addition to providing the formatted table data. We used the same rationale and procedure to select and fine-tune the models as outlined in the SQL query generation subsection.

## 4. RESULTS

### 4.1. Domain-specific Question-answering

Our experimental results provide an illuminating perspective on the efficacy of different LLMs paired with semantic search on domain-specific question-answering. This is an important capability because many of the questions asked of a marketing analytics assistant are informational in nature, and it is critical to give the user information that is accurate, relevant, thorough, clear, and concise. As depicted in Figure 2, utilizing GPT-4 without the aid of semantic search yielded suboptimal outcomes, scoring consistently lower than other methodologies across various evaluation metrics. However, a synergy between GPT-4 and semantic search bolstered the performance, with this combination achieving the highest scores across the board. Notably, the Llama-2-70b model exhibited metrics that were nearly on par with GPT-4, albeit slightly lagging. It is crucial to highlight the possibility of a methodological bias given that GPT-4 was responsible for the evaluations. Taking this into account, the results suggest that Llama-2-70b offers an economical yet effective alternative to GPT-4.

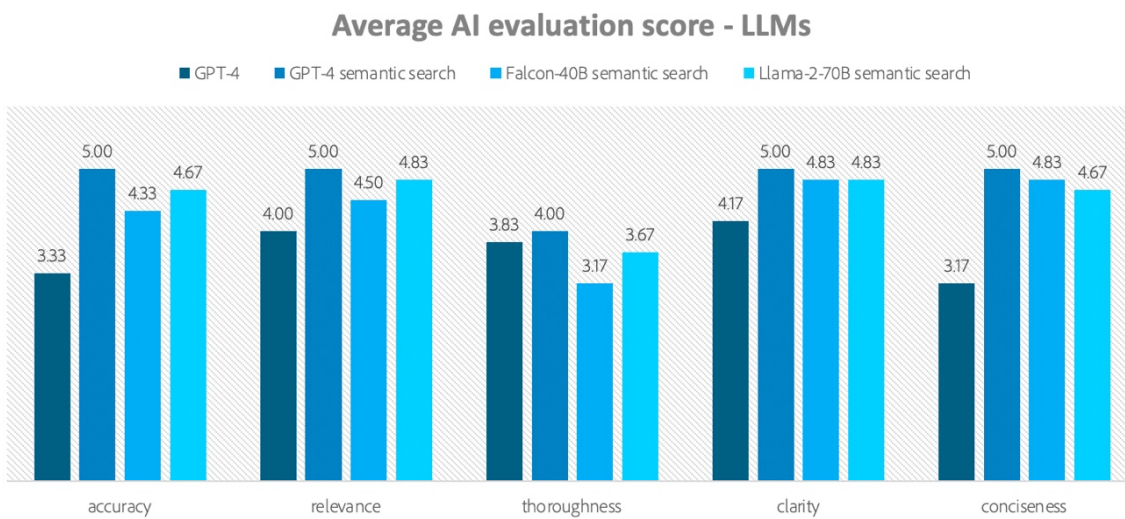

Figure 2. Average AI evaluation scores on the six marketing analytics related questions by different LLMs with text-embedding-ada-002 used as the text embedding model.

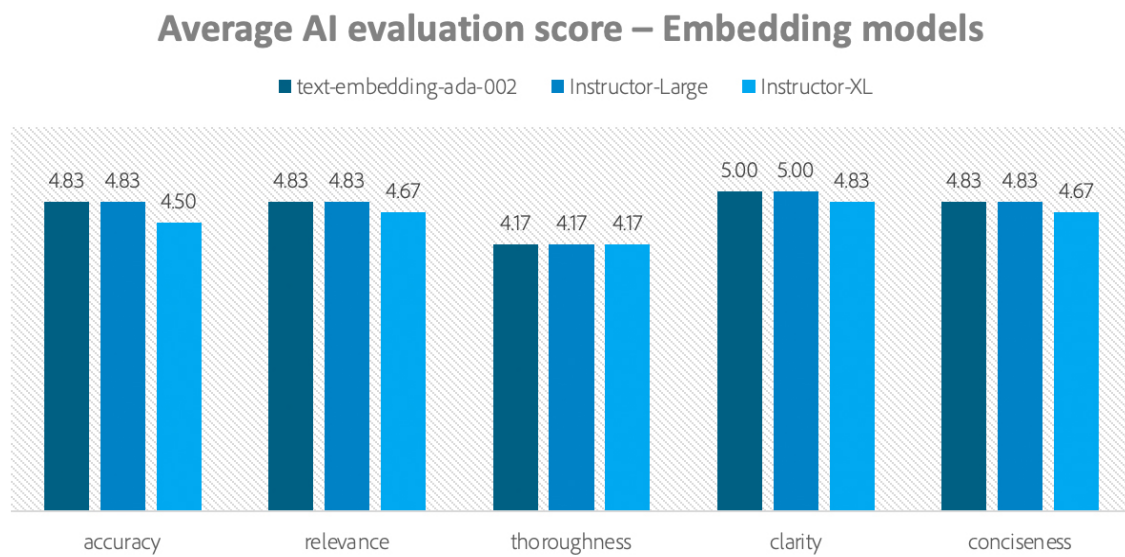

Figure 3. Average AI evaluation scores on the six marketing analytics related questions by different text embedding models paired with GPT-4.

The embedding model encodes the information used by semantic search to find the relevant context for use with the LLM and can potentially have a significant impact on overall performance. Our findings suggest that the performance of the LLMs on domain-specific question-answering is more sensitive to the LLM itself, as opposed to the embedding method. However, an interesting pattern emerged as shown in Figure 3. The INSTRUCTOR-LARGE model demonstrated performance metrics comparable to OpenAI's text-embedding-ada-002. In contrast, its more expansive counterpart, the INSTRUCTOR-XL, showed somewhat diminished results. Two potential explanations arise for this discrepancy. First, the expansive architecture of the INSTRUCTOR-XL may inadvertently capture noise, leading to overfitting and hindering generalization. Second, the increased dimensionality of embeddings from larger models introduces challenges in high-dimensional spaces. Distances between points, such as cosine similarities, may lose discriminative power due to the curse of dimensionality, impacting the efficiency of similarity-based retrieval.

Our findings demonstrate that open-source LLMs and text embedding models can be highly competitive with proprietary methods, like GPT-4 and text-embedding-ada-002. Specifically, the integration of Llama-2-70b and INSTRUCTOR-LARGE in semantic search endeavors appears promising, offering robust performance, with potential benefits related to cost, stability, privacy, and security.

### 4.2. SQL Query Generation

In evaluating various LLMs' ability to generate SQL queries, we systematically compared the LLMs' outputs against reference answers from our dataset, as illustrated in Table 1. Our results in Figure 4 revealed that GPT-4, utilizing a few-shot learning strategy, managed to achieve an accuracy rate of 64.5%. This is notable considering the intricate nature of SQL query formation. However, the performance of open-sourced pretrained models left much to be desired, struggling to surpass the 40% mark. Interestingly, post-fine-tuning provided a significant improvement in performance for both the Falcon-40b and Llama-2-13b models. Their accuracy rates increased from under 30% to over 80%, emphasizing the profound impact of fine-tuning on enhancing the quality of SQL query generation. Additionally, the Llama-2-70b model's preliminary performance was superior to both its Falcon counterpart and the smaller Llama variant, suggesting a tantalizing potential for even further enhancement after fine-tuning in the task of SQL query generations.

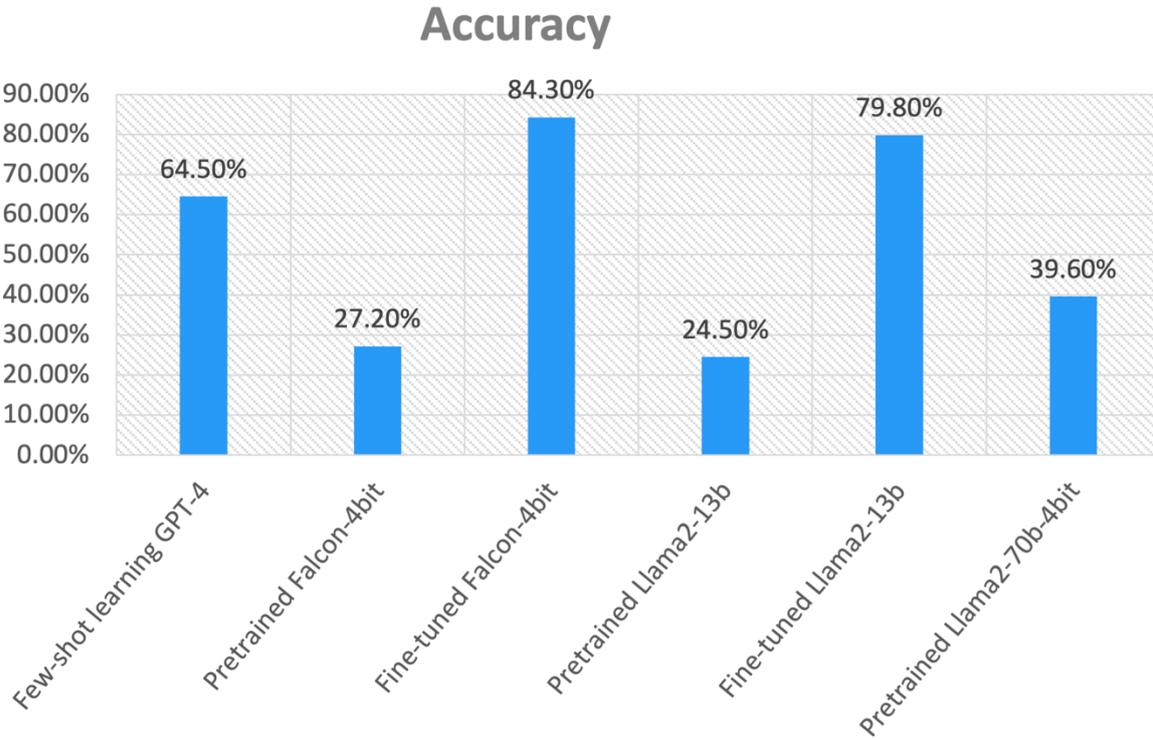

Figure 4. Accuracy of SQL query generations by different LLM approaches.

### 4.3. Tabular Data Analysis

During our analysis of tabular data comprehension, we benchmarked LLMs against programmatically generated ground truth answers, which converted numerical classifications into natural language explanations. The model accuracy in this comparison is detailed in Figure 5.

We found that methods using prompt engineering consistently reached about 70% accuracy, indicating that GPT-4 understands tabular structure but not numerical relationships. Changing data formats, like table rows to list or sentences, did not help. In contrast, GPT-4 paired with the Langchain pandas agent achieved a perfect 100% accuracy. The pretrained Falcon-40b and Llama-2-13b models initially struggled, barely achieving 10% accuracy, indicating difficulties in understanding both tabular structure and numerical relationships. Yet, post-fine-tuning, their performance drastically improved: Falcon-40b reached over 90% and Llama-2-13b over 80%, likely due to the LLMs' better understanding of csv file formats as well as basic numerical relationships. It is not clear what the relative contribution of each of these is to the overall improvement for each model. This increase in accuracy reaffirms fine-tuning's significance in tasks involving tabular data comprehension. Lastly, the Llama-2-70b model showed a robust initial accuracy of over 50%, hinting at its innate proficiency and the potential for even better results with further fine-tuning.

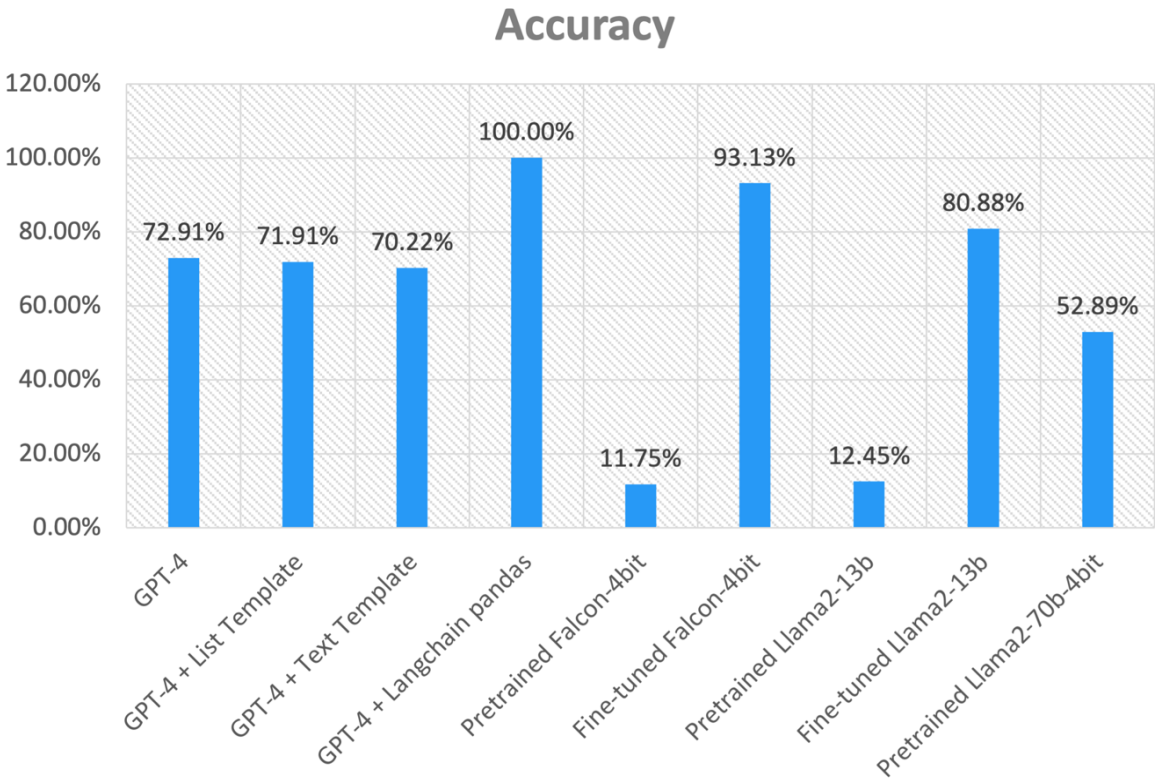

Figure 5. Accuracy of tabular data understanding by different LLM approaches.

## 5. DISCUSSION

Throughout this paper, we explored the capabilities and enhancement of various LLMs in domain-specific question-answering, SQL query generation, and tabular data analysis. Accurately performing these tasks are critical to effectively building an AI-based marketing analytics assistant. Here we discuss the implications of the results and other topics related to deployment.

For domain-specific question-answering, our experiments underscored the synergistic potential of combining GPT-4 with semantic search techniques. While GPT-4 alone exhibited suboptimal performance, its integration with semantic search techniques significantly bolstered its efficacy. Importantly, the Llama-2-70b model emerged as a promising alternative to GPT-4, especially when considering potential methodological biases caused by using GPT-4 to evaluate itself relative to other models. Similarly, the INSTRUCTOR-LARGE text embedding model showcased its potential as a robust alternative to OpenAI's offerings. Semantic search can be employed in any circumstance where text-based question-answering is important and there are one or more text documents related to the potential questions. In a marketing context, the user may ask questions about how a model works, what data is supported, and how to resolve errors that arise. In ecommerce, a customer may ask questions about different products on a website, while patients may have questions about medical procedures in a healthcare context. In all cases, providing accurate, thorough, clear, concise, and relevant answers are important, and semantic search greatly enhances an LLMs ability to achieve this.

Our exploration into SQL query generation revealed the utility of fine-tuning. While pretrained models like Falcon-40b and Llama-2-13b initially struggled with accuracy, fine-tuning results in a dramatic increase in accuracy, demonstrating the extent to which fine-tuning can improve model performance on relatively novel tasks. The base Llama-2-70b model showcased promising capabilities, suggesting that fine-tuning this model would result in accuracy significantly exceeding 84%. Importantly, this illustrates one of the weaknesses of relying on a proprietary model, like GPT-4. Since the specific architecture and weights of GPT-4 are not made public, it isn't possible to do fine-tuning on GPT-4. With fine-tuning, other models, including the relatively small Llama-2-13b model, are capable of significantly beating the performance of the current version of GPT-4 for SQL query generation. As it stands, our results suggest that larger open-sourced models, coupled with fine-tuning, may have the best performance on coding related tasks. In other words, specialized, as opposed to generalized, LLMs are likely to be most effective in solving coding related problems. Since the data used to evaluate this task is not specific to marketing, these results are expected to generalize to any discipline that uses SQL databases or similar technology, which is widely used across many industries.

With respect to tabular data analysis, our findings highlighted the superiority of fine-tuning over prompt engineering, especially with open-source LLMs. The fine-tuned Falcon-40b and Llama-2-13b models demonstrated remarkable improvements versus pretrained models with prompt engineering, underscoring the significant impact of fine-tuning on model accuracy. Again, it isn't possible to do fine-tuning on GPT-4, and as a result, open-sourced models are able to generally outperform GPT-4 for tabular analysis tasks. An exception to this is when GPT-4 is coupled to pandas agent, which is 100% accurate. This is notable because GPT-4 is actually the only LLM that performs well with pandas agent, likely because it performs better on the coding tasks that pandas agent orchestrates behind the scenes relative to other LLMs. Ultimately, GPT-4 is a very comprehensive LLM and performs remarkably well on a wide variety of tasks, even outside standard NLP tasks. However, other large open sourced LLMs can often outperform GPT-4 with fine-tuning. Fine-tuning open sourced LLMs to work effectively with pandas agent or even fine-tuning the Llama-2-70b model without the use of pandas agent may come close to, if not equal, the performance of GPT-4 coupled to pandas agent. Although our tabular analysis example was demonstrated on a very specific marketing use case, the results should generalize to other row-by-row table summarization tasks. For example, LLMs could be used to alert a supply chain professional when safety stock is below a predefined threshold.

Although GPT-4 does perform quite well in many circumstances, there are a variety of other reasons to use open-sourced models if the performance is at least reasonably close. Many people believe cost is one of the primary reasons to use open-source models. It is not entirely clear that this is true. Open-source models are free to use, but have significant computational, engineering, and maintenance costs to build and deploy them. At larger companies with lower computation costs (often due to bulk discounts on cloud computational resources) and extensive engineering resources to build and maintain these models, open-sourced models are likely less expensive. However, this may not be true for smaller companies. Proprietary models, like GPT-4, require paying a fee, but have much lower internal costs. Regardless, open-sourced models are always more flexible because their entire architecture is public. The additional control over the model is important, as the model may need fine-tuning and can be updated only when it experiences performance issues for its specific deployment. In contrast, proprietary models may be updated at any time by the provider, which can adversely impact stability and consistency. Moreover, open-sourced models have better security and privacy guarantees because no data needs to be transmitted to a third party. In this regard, open-sourced models are clearly superior, and their use may be justified even for smaller companies with more limited resources, making our research in semantic search and fine-tuning to improve open-sourced models all the more important.

## 6. FUTURE DIRECTIONS

Based on the results, there are several areas we are pursuing in terms of future research. Regarding domain-specific question-answering, LLMs tend to work quite well, especially when using semantic search techniques. Anecdotally, we have observed that LLMs paired with semantic search almost always return the correct answer if the answer is contained within one of the documents and the question is clear. It is important to address cases where these conditions are not satisfied. Having the LLM ask follow-up questions is one possible solution. Another alternative is to have the LLM rephrase the question to match one from a reference list, which are mapped to human generated answers. The information presented to the user is guaranteed to be correct, although it may not be helpful if it answers the wrong question. To avoid this, the rephrased question can be presented to the user to validate that it is the correct intent. In both cases, having a human in the loop is a critical component to effective deployment. Delving into the relative value of different strategies is beyond the scope of this paper but will be the subject of additional research.

Even with fine-tuning, LLMs were meaningfully less effective at writing SQL code in comparison to answering questions. We plan to study how a fine-tuned version of Llama-2-70b performs on this task, and this alone may be sufficient to resolve any lingering performance issues. However, it may also be necessary to switch our focus to writing API calls. While this paper focused on SQL query generation due to its universality and ease of access to labeled data, it is safer and more reliable to make API calls instead of SQL queries to retrieve data. APIs are typically designed to be simpler than SQL queries and can be programmatically checked much more easily. While we expect the relative performance of the various LLMs to be the same if applied to API calls, we think it is likely that the absolute performance will improve, but this will be verified in future work.

In this study, we showcased that LLMs can effectively summarize tables row-by-row with fine-tuning and/or the use of specialized agents, like the pandas agent. However, this is only one possible tabular analysis capability. Other potential use cases may include plotting relationships on a graph, analyzing trends, detecting anomalies, and otherwise summarizing the content of a table in a more sophisticated way. The most effective method to handle these other use cases will be the subject of future research. Of note, it is almost always possible to programmatically answer questions based on data stored in tables. In fact, the synthetic data set used for validation in the tabular analysis use case had programmatically generated answers that were guaranteed to be 100% accurate. To effectively deploy LLMs within marketing analytics tools, it is likely necessary to use the LLM primarily to understand a user's question and route the question to the right specialized agent. This is conceptually similar to what we were doing by using pandas agent, which is why this method performed so well. One drawback of this approach is that the dialogue may feel stilted. Though less important than accuracy, this is still a consideration in terms of user experience, and it may be useful for the LLM to rephrase the results in a way that is more concise and conversational. Both using LLMs for routing and pairing LLMs with other specialized agents will be the subject of future research. Beyond accuracy, one of the advantages of using LLMs primarily to route questions is that we can more easily put guardrails on the output to eliminate other harmful problems, such as bias.

All three of the capabilities discussed in this paper eventually need to be surfaced to a user, and user interface (UI) design is the subject of ongoing research. We considered surfacing the LLM via a popup window in the bottom right corner of our existing UI, which is often how customer service chatbots are surfaced on websites. However, this limits the types of insights that can be delivered. Tables and graphs will not be legible in a small popup window, and even lengthier text may be difficult to read. Therefore, the copilot should ideally have its own dedicated panel in the UI, where intermediate results can optionally be displayed separately for more technical users interested in understanding and modifying the chain of thoughts of the LLM and other specialized agents.

## 7. CONCLUSION

Each of the functionalities highlighted in this paper are critical to the deployment of a marketing analytics copilot. While not exhaustive, this work highlights a couple of key points about developing a marketing analytics assistant. First, it is useful to have specialized programs, often called agents, employing different LLMs to perform different tasks. Semantic search is most useful for providing LLMs with sufficient context to answer text-based questions, and GPT-4 could reasonably be used in these cases. However, SQL query generation and tabular analysis tasks benefit heavily from fine-tuning, which required open-source models with public architectures. In some cases, other programs, like pandas agent, may obviate the need for fine-tuning, but fine-tuning coupled with specialized agents are likely to perform best. Even in cases where proprietary models exhibit the best performance, open-source models have advantages in terms of flexibility, stability, privacy, and security and should be considered if their performance is close to that of proprietary models. These points extend beyond the application of LLMs to marketing analytics copilots, and our findings serve as a foundational guide for researchers and practitioners aiming to harness the power of LLMs in a wide variety of domains.

## REFERENCES

[1] A. Vaswani *et al.*, "Attention is all you need," *Advances in neural information processing systems*, vol. 30, 2017.

[2] J. Devlin, M.-W. Chang, K. Lee, and K. Toutanova, "BERT: Pre-training of Deep Bidirectional Transformers for Language Understanding," in *Proceedings of the 2019 Conference of the North American Chapter of the Association for Computational Linguistics: Human Language Technologies, NAACL-HLT 2019, Minneapolis, MN, USA, June 2-7, 2019, Volume 1 (Long and Short Papers)*, 2019, pp. 4171–4186.

[3] A. Radford and K. Narasimhan, "Improving Language Understanding by Generative Pre-Training," 2018.

[4] M. Lewis *et al.*, "BART: Denoising Sequence-to-Sequence Pre-training for Natural Language Generation, Translation, and Comprehension," in *Proceedings of the 58th Annual Meeting of the Association for Computational Linguistics*, Jul. 2020, pp. 7871–7880.

[5] L. Floridi and M. Chiriatti, "GPT-3: Its Nature, Scope, Limits, and Consequences," *Minds and Machines*, vol. 30, pp. 681–694, 2020.

[6] R. Thoppilan *et al.*, "LaMDA: Language Models for Dialog Applications," *ArXiv*, vol. abs/2201.08239, 2022.

[7] J. Geiping and T. Goldstein, Cramming: Training a Language Model on a Single GPU in One Day. 2022.

[8] T. Dao, D. Y. Fu, S. Ermon, A. Rudra, and C. R'e, "FlashAttention: Fast and Memory-Efficient Exact Attention with IO-Awareness," *ArXiv*, vol. abs/2205.14135, 2022.

[9] Y. Bai et al., "Constitutional AI: Harmlessness from AI Feedback," *ArXiv*, vol. abs/2212.08073, 2022.

[10] L. Ouyang *et al.*, "Training language models to follow instructions with human feedback," *ArXiv*, vol. abs/2203.02155, 2022.

[11] K. Shuster *et al.*, "BlenderBot 3: a deployed conversational agent that continually learns to responsibly engage," *ArXiv*, vol. abs/2208.03188, 2022.

[12] A. Glaese *et al.*, "Improving alignment of dialogue agents via targeted human judgements," *ArXiv*, vol. abs/2209.14375, 2022.

[13] T. L. Scao *et al.*, "BLOOM: A 176B-Parameter Open-Access Multilingual Language Model," *ArXiv*, vol. abs/2211.05100, 2022.

[14] S. Peng, E. Kalliamvakou, P. Cihon, and M. Demirer, "The Impact of AI on Developer Productivity: Evidence from GitHub Copilot," *ArXiv*, vol. abs/2302.06590, 2023.

[15] Y. Mehdi, "Announcing Microsoft Copilot, your everyday AI companion," *The Official Microsoft Blog*, 2023.

[16] N. McKenna, T. Li, L. Cheng, M. J. Hosseini, M. Johnson, and M. Steedman, "Sources of Hallucination by Large Language Models on Inference Tasks," *ArXiv*, vol. abs/2305.14552, 2023.

[17] Y. Zhang *et al.*, "Siren's Song in the AI Ocean: A Survey on Hallucination in Large Language Models," *ArXiv*, vol. abs/2309.01219, 2023.

[18] Y. Bang *et al.*, "A Multitask, Multilingual, Multimodal Evaluation of ChatGPT on Reasoning, Hallucination, and Interactivity," *ArXiv*, vol. abs/2302.04023, 2023.

[19] N. H. Borden, "The concept of the marketing mix," *Journal of advertising research*, vol. 4, no. 2, pp. 2–7, 1964.

[20] E. J. McCarthy, "Basic marketing: a managerial approach," *(No Title)*, 1978.

[21] D. Chan and M. Perry, "Challenges and opportunities in media mix modeling," 2017.

[22] H. Li and P. Kannan, "Attributing conversions in a multichannel online marketing environment: An empirical model and a field experiment," *Journal of marketing research*, vol. 51, no. 1, pp. 40–56, 2014.

[23] X. Shao and L. Li, "Data-driven multi-touch attribution models," in Proceedings of the 17th ACM SIGKDD international conference on Knowledge discovery and data mining, 2011, pp. 258–264.

[24] B. Dalessandro, C. Perlich, O. Stitelman, and F. Provost, "Causally motivated attribution for online advertising," in *Proceedings of the sixth international workshop on data mining for online advertising and internet economy*, 2012, pp. 1–9.

[25] W. Ji, X. Wang, and D. Zhang, "A probabilistic multi-touch attribution model for online advertising," in *Proceedings of the 25th acm international on conference on information and knowledge management*, 2016, pp. 1373–1382.

[26] Y. Zhang, Y. Wei, and J. Ren, "Multi-touch attribution in online advertising with survival theory," in *2014 IEEE International Conference on Data Mining*, 2014, pp. 687–696.

[27] W. Ji and X. Wang, "Additional multi-touch attribution for online advertising," in *Proceedings of the AAAI Conference on Artificial Intelligence*, 2017, vol. 31, no. 1.

[28] V. Abhishek, P. Fader, and K. Hosanagar, "Media exposure through the funnel: A model of multi-stage attribution," *Available at SSRN 2158421*, 2012.

[29] N. Li, S. K. Arava, C. Dong, Z. Yan, and A. Pani, "Deep Neural Net with Attention for Multi-channel Multi-touch Attribution," *ArXiv*, vol. abs/1809.02230, 2018.

[30] J. Tao, Q. Chen, J. W. Snyder, A. S. Kumar, A. Meisami, and L. Xue, "A Graphical Point Process Framework for Understanding Removal Effects in Multi-Touch Attribution," *ArXiv*, vol. abs/2302.06075, 2023.

[31] A. Calsavara and G. Schmidt, "Semantic Search Engines," in *Advanced Distributed Systems*, 2004, pp. 145–157.

[32] A. Léger, L. J. B. Nixon, P. Shvaiko, and J. Charlet, "Semantic Web applications: Fields and Business cases. The Industry challenges the research.," in *Industrial Applications of Semantic Web*, 2005, pp. 27–46.

[33] F. Burkhardt, J. A. Gulla, J. Liu, C. Weiss, and J. Zhou, "Semi-Automatic Ontology Engineering in Business Applications," in *38. Jahrestagung der Gesellschaft für Informatik, Beherrschbare Systeme - dank Informatik, INFORMATIK 2008, Munich, Germany, September 8-13, 2008, Band 2*, 2008, vol. P-134, pp. 688–693.

[34] S. Sharma, S. Mahajan, and V. Rana, "A semantic framework for ecommerce search engine optimization," *International Journal of Information Technology*, vol. 11, no. 1, pp. 31–36, Jul. 2018.

[35] P. Lu *et al.*, Chameleon: Plug-and-Play Compositional Reasoning with Large Language Models. 2023.

[36] M. Sallam, "The Utility of ChatGPT as an Example of Large Language Models in Healthcare Education, Research and Practice: Systematic Review on the Future Perspectives and Potential Limitations," *medRxiv*, 2023.

[37] A. Drozdov *et al.*, *Compositional Semantic Parsing with Large Language Models*. 2022.

[38] T. Brown *et al.*, "Language Models are Few-Shot Learners," in *Advances in Neural Information Processing Systems*, 2020, vol. 33, pp. 1877–1901.

[39] J. Yosinski, J. Clune, Y. Bengio, and H. Lipson, "How Transferable Are Features in Deep Neural Networks?," in *Proceedings of the 27th International Conference on Neural Information Processing Systems - Volume 2*, 2014, pp. 3320–3328.

[40] P. Yin and G. Neubig, "A Syntactic Neural Model for General-Purpose Code Generation," in *Proceedings of the 55th Annual Meeting of the Association for Computational Linguistics (Volume 1: Long Papers)*, Jul. 2017, pp. 440–450.

[41] C. D. Manning, P. Raghavan, and H. Schütze, *Introduction to information retrieval*. Cambridge university press, 2008.

[42] S. Iyer, I. Konstas, A. Cheung, J. Krishnamurthy, and L. Zettlemoyer, "Learning a neural semantic parser from user feedback," in *Proceedings of the 55th Annual Meeting of the Association for Computational Linguistics (Volume 1: Long Papers)*, 2017, pp. 963–973.

[43] V. Zhong, C. Xiong, and R. Socher, "Seq2SQL: Generating Structured Queries from Natural Language using Reinforcement Learning," *CoRR*, vol. abs/1709.00103, 2017.

[44] R. Sun et al., *SQL-PaLM: Improved Large Language Model Adaptation for Text-to-SQL*. 2023.

[45] V. Câmara, R. Mendonca-Neto, A. Silva, and L. Cordovil-Jr, "DBVinci – towards the Usage of GPT Engine for Processing SQL Queries," in *Proceedings of the 29th Brazilian Symposium on Multimedia and the Web*, 2023, pp. 91–95.

[46] Z. Gu *et al.*, *Interleaving Pre-Trained Language Models and Large Language Models for Zero-Shot NL2SQL Generation*. 2023.

[47] Q. Liu et al., "TAPEX: Table Pre-training via Learning a Neural SQL Executor," in *International Conference on Learning Representations*, 2022.

[48] L. Zha *et al.*, *TableGPT: Towards Unifying Tables, Nature Language and Commands into One GPT*. 2023.

[49] H. Su *et al.*, "One Embedder, Any Task: Instruction-Finetuned Text Embeddings," 2022.

[50] E. Almazrouei et al., "Falcon-40B: an open large language model with state-of-the-art performance," 2023.

[51] H. Touvron et al., *Llama 2: Open Foundation and Fine-Tuned Chat Models*. 2023.

[52] Y. Liu, D. Iter, Y. Xu, S. Wang, R. Xu, and C. Zhu, *G-Eval: NLG Evaluation using GPT-4 with Better Human Alignment*. 2023.